\documentclass[letterpaper, 10 pt, conference]{ieeeconf}  % Comment this line out if you need a4paper
\usepackage{graphicx}
\usepackage{xcolor}

\usepackage{cite}

\usepackage{amsmath}

\IEEEoverridecommandlockouts                              % This command is only needed if 
\title{\LARGE \bf
Unifying Sidewinding and Rolling: A Wave-Based Framework for Self-Righting in Elongated Limbless and Multi-Legged Robots
}

\author{Hangjun Liu, Jiarui Geng, Jinxuan Ding, Gengzhi He, Xiyuan Wang, Melisa Arukgoda, \\
Joe DiGennaro, George Ubertalli, Grigoriy Blekherman, Baxi Chong}

\begin{document}

\maketitle
\thispagestyle{empty}
\pagestyle{empty}

%%%%%%%%%%%%%%%%%%%%%%%%%%%%%%%%%%%%%%%%%%%%%%%%%%%%%%%%%%%%%%%%%%%%%%%%%%%%%%%%
\begin{abstract}

Centipede-like robots offer unique locomotion advantages due to their small cross-sectional area for accessing confined spaces. Their redundant legs enhance locomotion robustness, enabling reliable operation in cluttered environments such as search-and-rescue and pipe inspection. However, elongated robots are particularly vulnerable to tipping over when climbing over large obstacles. Reliable self-righting is therefore essential for field deployment. However, self-righting strategies for elongate, multi-legged systems remain poorly understood. In this study, we conduct a comparative biomechanics and robophysical investigation to address three key questions: (1) What self-righting strategies are effective for elongate, many-legged systems? (2) How should these strategies depend on morphological parameters such as leg length and leg number? (3) Is there a morphological limit beyond which reliable self-righting becomes infeasible? We first compare two biological exemplars with distinct limb morphologies: \textit{Scolopendra Subspinipes} (short legs) and \textit{Scutigera Coleoptrata} (house centipedes with long legs). We observe that \textit{Scolopendra Subspinipes} can reliably self-right both during aerial phases (analogous to cat-like reorientation) and through ground-assisted self-righting, whereas house centipedes rely predominantly on aerial reorientation and experience significant difficulty generating effective self-righting torques during ground contact. Motivated by these observations, we construct a parameterized space of bio-inspired self-righting strategies and develop an elongate robot with adjustable leg lengths. Systematic experiments reveal that increasing leg length necessitates a shift in control strategy to prevent torque over-concentration in mid-body actuators. Moreover, we experientially identify a critical limb-length threshold above which robust self-righting strategies can be challenging. These results establish morphology–strategy coupling principles for self-righting in elongate robots and provide design guidelines for centipede-like systems operating in uncertain terrain.
\end{abstract}

%%%%%%%%%%%%%%%%%%%%%%%%%%%%%%%%%%%%%%%%%%%%%%%%%%%%%%%%%%%%%%%%%%%%%%%%%%%%%%%%
\section{INTRODUCTION}

Legged robots demonstrate superior mobility over wheeled platforms when traversing uneven terrain, partially because they can actively lift their limbs to step over obstacles and adapt to complex ground geometry~\cite{sun_rhex}. However, lifting legs reduces static stability and increases susceptibility to tipping, particularly in cluttered or unpredictable environments. Despite advances in control and algorithms, loss of balance and falling over remain typical failure modes in many real-world deployments \cite{Gui_MPCSR}.
% [CITE]
This vulnerability of flipping in practical deployments makes self-righting (the ability to recover from an overturned state) a critical capability for ensuring robust operation in unstructured terrain \cite{casarez_thesis_tail}. Over the past decades, prior work has demonstrated successful self-righting strategies across various robotic platforms, including passive self-righting \cite{kovac2009passiveJumper,Tsukagoshi_jumpbot}, appendage-assisted mechanisms (e.g., wings ~\cite{li_wingroach} or tails ~\cite{casarez_tail}), dynamic inertial maneuvers~\cite{Chang_aerialSR, JKYim_salto, libby_tail, yamafuji_cat}, and body-driven self-righting strategies~\cite{peng2017motion, Tan_reconfig,erikSR}.

% when confronted with unconventional morphologies, it becomes unclear whether existing strategies remain applicable.
Despite these successes, existing self-righting strategies are largely developed in a platform-specific, \textit{ad hoc} manner partially because identifying a single reliable recovery maneuver is often sufficient for a given robot. However, this case-by-case approach lacks the ability to be generalized across different robotic platforms. Thus, it becomes unclear whether those platform-specific self-righting strategies remain feasible for robots with new/unconventional morphologies.
For multi-legged elongated robots \cite{he_tactile_elongate}, fundamental self-righting questions remain less explored. For example, 
% (1) What mechanisms enable reliable self-righting? 
(1) How should self-righting strategies vary as a function of morphology, such as leg number or leg length? and (2) What are some of the morphological limits (e.g. leg length and leg number) below which reliable self-righting methods can be found for elongated, multi-legged robots?
% Is there a morphological limit beyond which reliable self-righting becomes infeasible?

\begin{figure}[t]
    \centering
    \includegraphics[width=\linewidth]{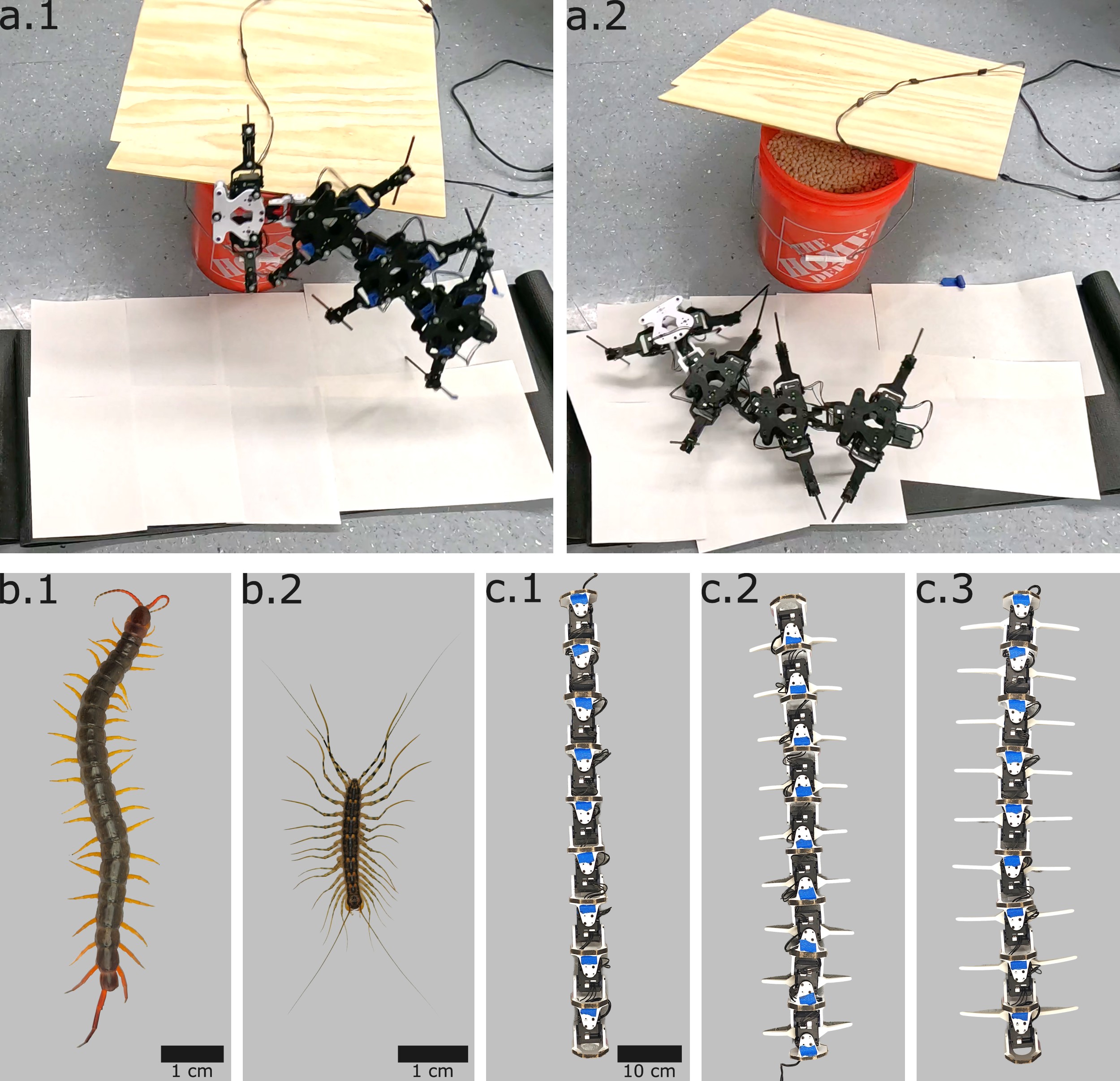}
    \vspace*{-5mm}
    \caption{\textbf{Centipedes and centipede-inspired robot.} 
    \textbf{(A)} Example of a centipede robot necessitating self-righting. (a.1) A centipede robot fell over from a elevated platform. (a.2) The centipede robot flipped upside down after falling from the platform. 
    \textbf{(B)} Biological centipedes with distinct limb morphologies: (b.1) \textit{Scolopendra subspinipes}, characterized by relatively short limbs, and (b.2) a house centipede, characterized by elongated limbs. 
    \textbf{(C)} Centipede-inspired robotic platform with adjustable limb length: (c.1) no limbs, (c.2) short limbs, and (c.3) long limbs.}
    \label{fig:fig1}
\end{figure}%

% One might question how frequently multi-legged robots are expected to overturn. In practice, the likelihood is non-negligible in realistic operating environments. 
% On relatively flat terrain, the low center of mass characteristic of elongated, multi-legged platforms provides inherent static stability, reducing the probability of spontaneous overturning.

Although multi-legged elongated robots possess a low center of mass (CoM) and multiple leg contacts that provide strong static stability on flat terrain, they experience frequent overturning in realistic operating environments. In practical deployments (such as climbing slopes and traversing rubble), interaction with large obstacles can often cause robot configurations that substantially increase tipping risk (Fig.\ref{fig:fig1} a.1 and a.2). Importantly, because the center of mass remains close to the ground in both upright and inverted states, the same morphological feature that enhances resistance to tipping on flat terrain also making self-righting mechanically challenging. As a result, developing reliable self-righting strategies for multi-legged elongated robots presents a significant mechanical and control challenge. With the recent advancement of multi-legged systems designed to operate in confined and cluttered environments \cite{KohCentipedeRF,itoRubbles,flores_steering}, establishing generalizable self-righting principles for a broad class of elongated platforms has become increasingly critical.

%This self-righting ability is particularly critical for centipede robots because they can be frequently actuated into statically unstable configurations during locomotion, which makes them vulnerable to tipping on uneven terrains. In addition, the low profile of centipede robots make their center of mass close to the ground, and their multi-legged morphology, which adds an additional energy barrier that the robot has to overcome, make self-righting for centipede robots particularly challenging. Other robotic platforms, from quadrupeds to snake-like robots, are also vulnerable to tip-over on uneven terrains, demonstrating the need for a systematic understanding of self-righting mechanisms for robust locomotion in complex environments. However, existing self-righting strategies have been developed largely on an ad-hoc basis, tailored to specific robot designs and lack the ability to be generalized to robotic platforms beyond their own type XX EXAMPLES FROM OTHER PAPERS XX.

%These robot-specific self-righting approaches have left a significant gap in our understanding of self-righting as a broader locomotion phenomenon. A generalized framework is beneficial for us to identify the mechanical or morphological features of a robot that lead to successful self-righting and the strategies that are applicable across multiple types of robotic platforms. A systematic understanding of the principles of robot self-righting will provide valuable guidelines in both the hardware and the gait algorithm aspects to robot designs for robust locomotion in complex environments.

In this paper, we conduct a comparative biomechanics and robophysical study. We first record the self-righting behaviors of \textit{Scolopendra subspinipes} and house centipedes (\textit{Scutigera coleoptrata}) dropped from different heights, revealing distinct recovery strategies employed by each species under varying falling conditions. Motivated by these observations, we introduce a general kinematic framework to unify previously documented snake rolling and sidewinding gaits. We test this framework using a robophysical model equipped with varying numbers of pairs of static (non-actuated) legs of different lengths. Systematic experiments quantify self-righting probability and lateral displacement across morphological and control parameter spaces. Specifically, we analyze how morphology (limb length and limb number) and control parameters (body wave amplitude and spatial frequency) jointly influence self-righting success, and we identify morphological boundaries beyond which reliable self-righting becomes substantially more difficult. Finally, we demonstrate that while limbs can impede certain self-righting maneuvers, they can also stabilize body rolling and enhance sidewinding performance, achieving up to 0.8 body lengths per cycle (more than twice previously reported sidewinding speeds for comparable systems). Together, these results establish quantitative morphology–strategy coupling principles and provide design guidelines for centipede-like elongated robots operating in complex terrain.

\section{SELF-RIGHTING IN CENTIPEDES}

\begin{figure}
    \centering
    \includegraphics[width=\linewidth]{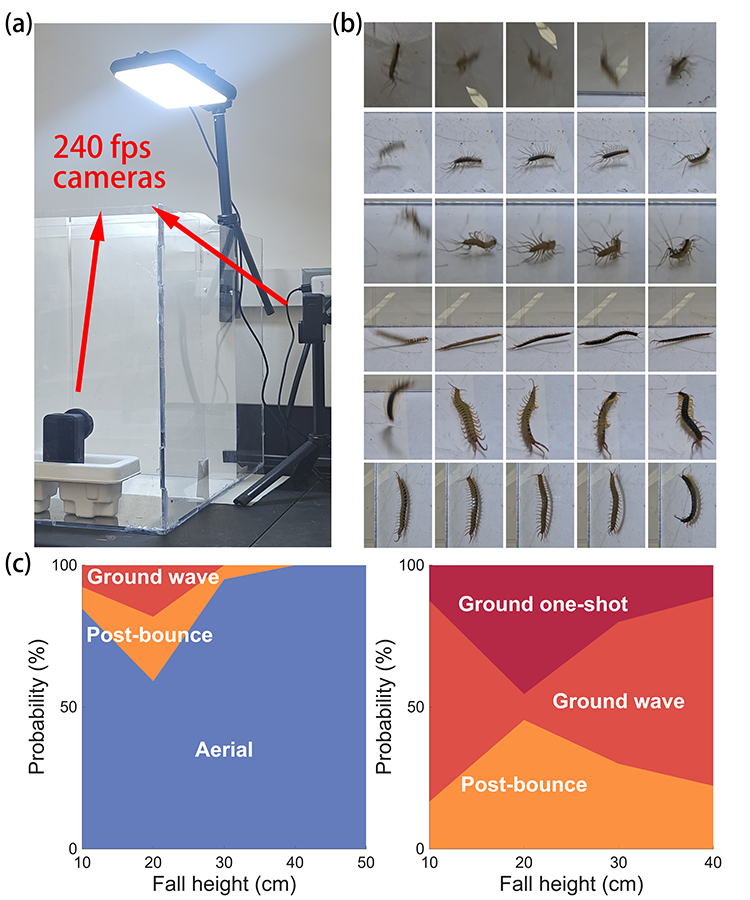}
    \vspace*{-5mm}
    \caption{\textbf{Centipede drop-test setup, representative self-righting modes, and height-dependent mode probabilities.}
    (a) Controlled drop arena and imaging configuration using two synchronized 240 fps high-speed cameras (front and side views).
    (b) Representative image sequences for modes observed in each species. House centipedes exhibit \textit{aerial righting}, \textit{post-bounce righting}, and \textit{ground wave righting}. \textit{S.~subspinipes} exhibits \textit{post-bounce righting}, \textit{ground wave righting}, and \textit{ground one-shot righting} (often accompanied by a pronounced C-shaped body posture).
    (c) Mode probability as a function of drop height for house centipedes (left) and \textit{S.~subspinipes} (right), shown as stacked distributions (each height: $\geq$10 trials; three individuals per species).}
    \label{fig:fig2}
\end{figure}%

\subsection{Experiment set-up}

To probe how morphology shapes self-righting strategy in elongate, many-legged systems, we conduct animal experiments on two biological exemplars with different limb size: house centipedes (\textit{Scutigera coleoptrata}, “long-leg” morphology 0.7 Body Length (BL)) and \textit{Scolopendra subspinipes} (“short-leg” morphology 0.13 BL). Notably, our goal is not to exhaustively model neuromuscular control, but to extract a compact, reproducible description of self-righting principles that can inform robophysical strategy parameterization. 

%We quantify how falling condition (drop height) redistributes the probability of distinct self-righting modes and how the characteristic righting timescales depend on both morphology and mode.

Experiments are conducted in a controlled drop arena (Fig. 2a). We record each trial using two high-speed cameras (240 fps) placed to capture a front view and a side view of the animal. Dual-view imaging enables reliable identification of releasing and dynamics of behaviors before landing and post-landing. We use two cameras to minimize ambiguity caused by occlusion or foreshortening in any single view. 
All time scales reported below are computed from annotated frame indices divided by the camera frame rate.

House centipedes are dropped from five heights (10, 20, 30, 40, and 50 cm). \textit{S. subspinipes} were dropped from four heights (10, 20, 30, and 40 cm). For each species and height, we performed at least 10 trials. For each species, we tested three individuals, and the sequence of trials were randomized across individuals and heights to reduce fatigue. We did not test \textit{S. subspinipes} at 50 cm to avoid potential injury. Each trial start from a released, inverted state and ended when the animal either returned to a stable upright posture or failed to do so within the observation window.

\begin{figure}
    \centering
    \includegraphics[width=\linewidth]{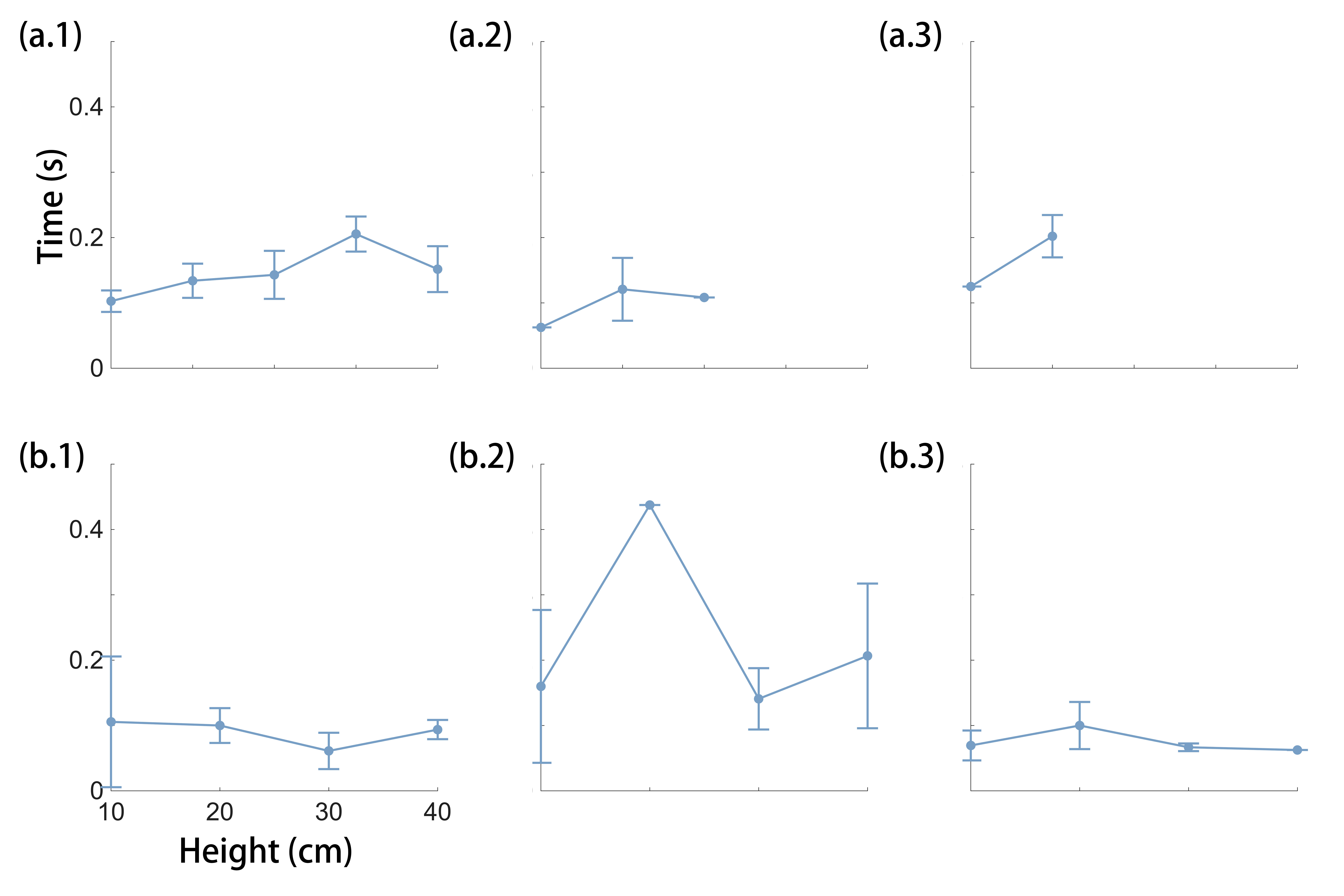}
    \caption{\textbf{Righting time versus drop height for each self-righting mode.}
    Top row: house centipedes, from left to right: aerial, post-bounce, and ground wave righting.
    Bottom row: \textit{S.~subspinipes}, from left to right: post-bounce, ground wave, and ground one-shot righting (aerial righting was not observed for \textit{S.~subspinipes} in our tests).
    Points show mean righting time; error bars indicate mean $\pm$ \textit{standard error of the mean}. Modes not observed at a given height are omitted events.}
    \label{fig:fig3}
\end{figure}%
For each trial we annotate three key events: (i) $t_{impact}$, the first frame of ground contact; (ii) $t_{start}$, the first frame showing a clear mode-specific reorientation attempt; and (iii) $t_{end}$, the first frame in which the animal completed an upright posture and remained stable.  

We classify self-righting into four distinct modes based on the timing and mechanism of body reorientation relative to impact and ground contact (Fig.~\ref{fig:fig2}b):
\begin{enumerate}
    \item \textbf{Aerial righting:} Reorientation is completed before the first ground contact. The righting duration is defined as: $t_{end} - t_{start}$ 
    \item \textbf{Post-bounce righting:} The centipede first contacts the ground in an inverted (upside-down) state. The ground-reaction impulse induces a rebound, during which rapid reorientation is completed before the second ground contact. The righting duration is defined as $t_{\mathrm{end}} - t_{\mathrm{impact}}$.
    \item \textbf{Ground wave righting:} Centipede fails to self-right in the air and lands upside down post-bounce. Reorientation occurs on the ground, beginning at one end of the body and propagating longitudinally as a traveling wave. The righting duration is defined as $t_{end} - t_{start}$ 
    \item \textbf{Ground one-shot righting:} Centipede fails to self-right in the air and lands upside down post-bounce. Reorientation occurs on the ground through a single, rapid, whole-body maneuver without a clear propagating wave pattern. Righting duration: $t_{end} - t_{start}$ 
\end{enumerate}

\subsection{Self-righting strategies in different centipedes}

We report (a) the probability of each mode as a function of the fall height (Fig. 2c) and (b) the mean righting time as a function of height for each mode (Fig. 3). Error bars indicate mean ± standard error of the mean (SEM) unless otherwise noted.

Across heights, house centipedes predominantly self-right via aerial reorientation, with post-bounce and ground wave righting occurring with lower probability depending on the height (Fig. 2c, left). In contrast, \textit{S. subspinipes} rarely exhibit aerial righting in our tests; its self-righting outcomes redistribute among off-bump, ground wave, and ground one-shot modes as the height increases (Fig. 2c, right).

Righting time exhibits clear mode dependence (Fig.~\ref{fig:fig3}). For house centipedes (top row), aerial, post-bounce, and ground-wave modes occupy distinct timescales. 
% {\color{red}GB: It doesn't look that different to me, across the top row at least}. 
In particular, ground-based righting requires substantially longer time compared to aerial or post-bounce reorientation, which is consistent with their preference to self-right in the air if we release them at a higher distance. At 10~cm, the mean ground-wave righting time was $\mu_{\mathrm{ground\text{-}wave}}=0.125$~s, compared with $\mu_{\mathrm{aerial}}=0.103$~s and $\mu_{\mathrm{post\text{-}bounce}}=0.0625$~s. At 20~cm, the mean ground-wave righting time was $\mu_{\mathrm{ground\text{-}wave}}=0.202$~s, compared with $\mu_{\mathrm{aerial}}=0.134$~s and $\mu_{\mathrm{post\text{-}bounce}}=0.121$~s. For \textit{S.~subspinipes} (bottom row), the characteristic righting time of ground one-shot maneuvers is comparable to that of post-bounce righting across the tested drop heights (Fig.~\ref{fig:fig3}). Instead of interpreting the longer duration of ground-wave righting purely as “slower,” we view it as a mechanically distinct, controllable recovery primitive: the propagating-wave deformation redistributes contact forces and righting torque along the body over time, potentially reducing peak torque/power demands compared with a single impulsive maneuver. 

Moreover, unlike aerial or post-bounce righting, which depend strongly on fall height and impact/rebound conditions, ground-based righting can be initiated deliberately after the locomotor has already come to rest, making it a more transferable target for robophysical modeling and morphological scaling. We identify two distinct ground-based self-righting strategies in centipedes: a rapid one-shot maneuver and a slower propagating-wave maneuver. Notably, the one-shot strategy is observed only in the short-legged species. These observations suggest a strong coupling between morphology and self-righting strategy, which we further quantify through robophysical experiments in the following sections.

\begin{figure}
    \centering
    \includegraphics[width=\linewidth]{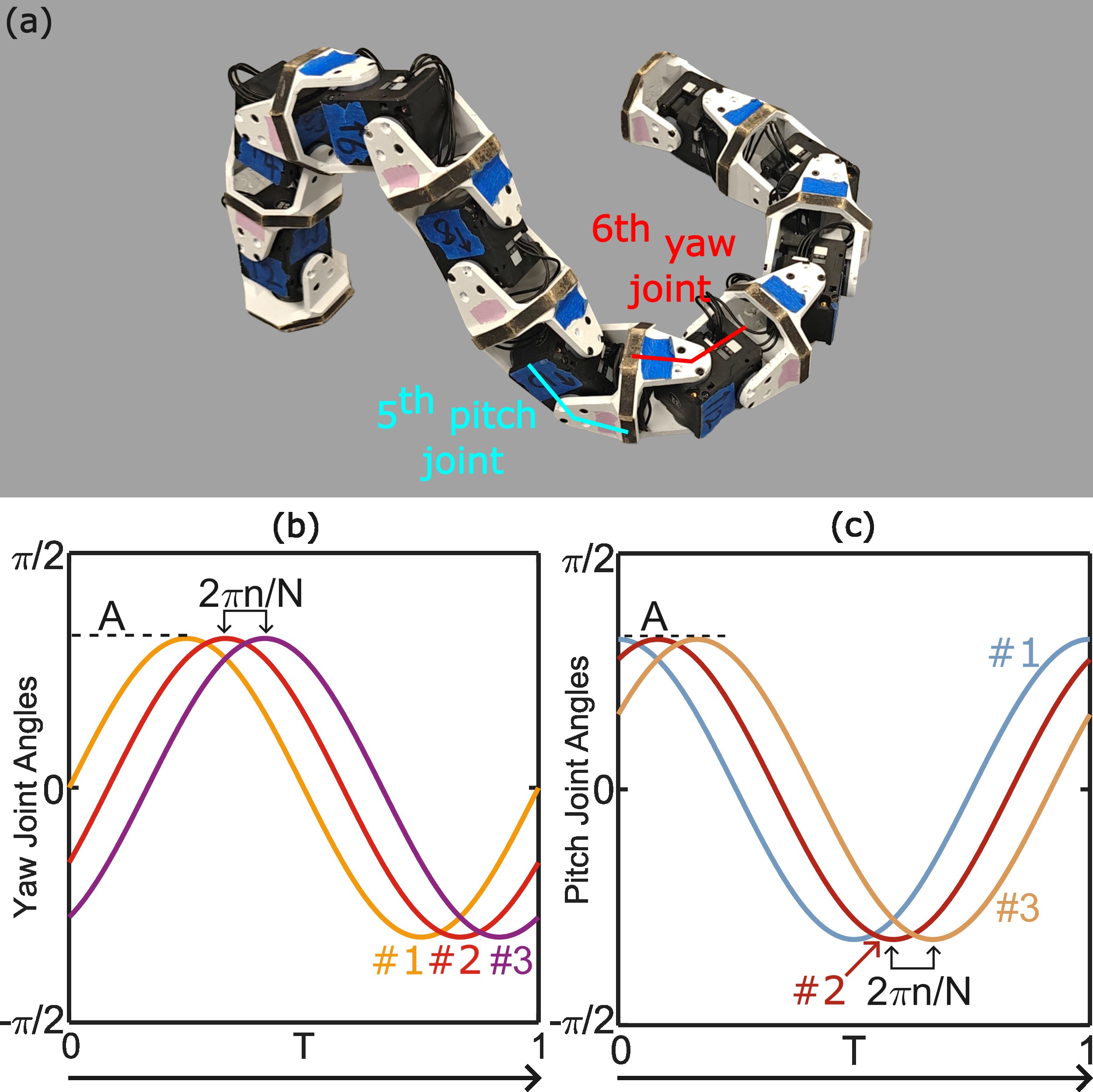}
    \caption{\textbf{Robot self-righting gait} 
    \textbf{(a)} The limbless robophysical model consists of a series of alternating yaw and pitch joints, with a yaw and a pitch joint labeled. 
    \textbf{(b)} The sinusoidal signals sent to the (\textit{left}) yaw joints and (\textit{right}) pitch joints. The joint amplitude is denoted by $A$ and the phase shift between adjacent yaw (\textit{and pitch}) joints is denoted by $2\pi n/N$.} %{\color{red} GB: is this figure in the right page? Robot design doesn't start until next page.}
    \label{fig:robo}
\end{figure}%

\section{THE ROBOPHYSICAL MODEL}

\begin{figure}
    \centering
    \includegraphics[width=\linewidth]{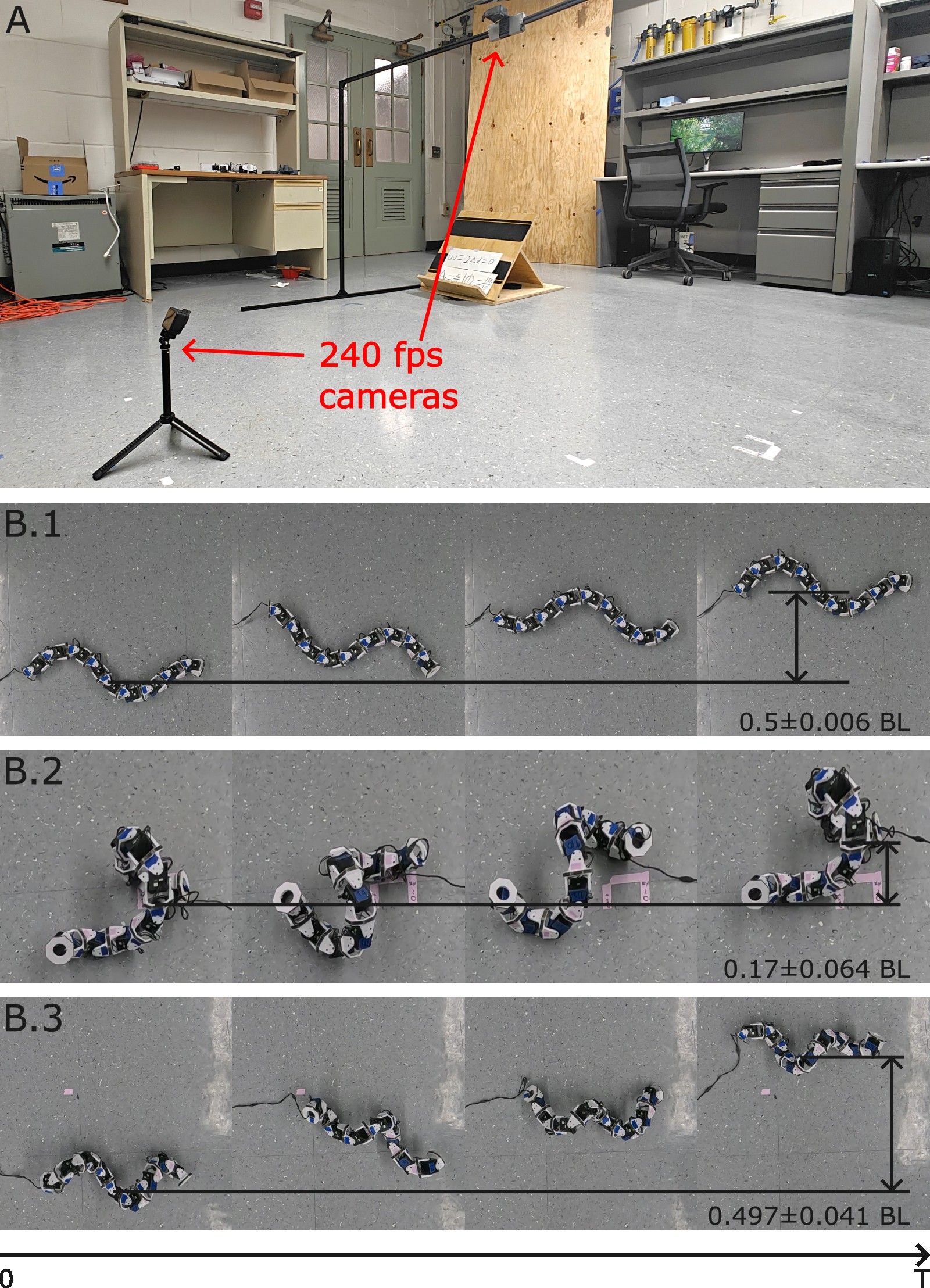}
    \caption{\textbf{Experimental setup and representative locomotion modes.} 
    \textbf{(A)} Two mounted 240 fps cameras capturing robot motion from top and side perspectives for kinematic analysis. 
    \textbf{(B)} Time-lapse snapshots of the limbless robot from $t=0$ to $t=T$: 
    (B.1) no-spin sidewinding (pure lateral translation without axial rotation), 
    (B.2) in-place spinning (axial rotation with minimal net translation), and 
    (B.3) sidewinding spin (coupled axial rotation and lateral translation).}
    \label{fig:fig4}
\end{figure}%

\subsection{Robot Designs}

We build a robophysical model (see Fig.~\ref{fig:robo} (a)) to reconstruct the self-righting behaviors and test their underlying principles. 
Our robot is constructed by a chain of nine Dynamixel 2XL430-W250T biaxial servomotors with each motor containing two output shafts with their axes of rotation orthogonal to each other. The output shafts (joints) are labeled from 1 to 18, where odd number joints are yaw (lateral) joints and even number joints are pitch (vertical) joints, The $i^{th}$ motor ($i \in \{1,9\}$) contains the output shafts labled by $(2i-1)$ and $(2i)$ (for example, motor 1 houses shafts 1 and 2, motor 2 houses shafts 3 and 4). 
The consecutive shafts of neighboring motors are connected by 3D-printed links (for example, shaft 2 to shaft 3 and shaft 4 to shaft 5) and construct the main body of the robot, with identical head and tail links mounted to joint 18 and joint 1, respectively. The robot is powered by a 12V DC power supply.
% Our model is inspired by the design of modular snake robots \cite{chosetSnake}, consisting of a chain of 9 total servomotors where the axis of rotation for each servomotor is offset 90 degrees from the previous servo's axis of rotation, resulting in a chain of servos providing alternating lateral and vertical degrees of freedom (see Fig.~\ref{fig:fig1} c.1). 
% This setup allows for the implementation of the two travelling waves prescribed by our model. The 3D-printed joints between servos are labelled by \textit{i} from 1 to 9, where joints with odd \textit{i} are vertical and joints with even \textit{i} are lateral. 

% The self-righting strategy we propose is driven exclusively by body undulation in the lateral and vertical planes. However, the presence of legs can introduce additional resistance to rolling~\cite{khazoom2022humanoid} and affect the self-righting effectiveness. To test the effect of legs, we compare two robot morphologies: with and without the legs. Specifically, we prepared static (non-actuated) limbs, made of laser-cut acrylic  attached to the robophysical body via 3D printed brackets (see Fig.~\ref{fig:fig4}c).

% Red markers were wrapped around each end of the model for position tracking, and green markers were placed on the top and yellow on the bottom for orientation tracking. The robot was powered by a constant voltage  of 11.3 V from a DC power supply. 

In addition to the limbless morphology that makes our robophysical model a snake-like robot, we design three sets of legs to be assembled at the bottom of the robot to empirically determine the self-righting success rate and the lateral displacement with different leg lengths and different numbers of legs. 
All three leg configurations are designed based on morphological parameters measured from the \textit{Scolopendra subspinipes} used in this study. 
The short-leg condition is derived from the ratio between the animal’s shortest leg length and total body length (approximately $1:13.8$). 
The medium-leg condition is based on the ratio between the average leg length and total body length (approximately $1:11.8$). 
The long-leg condition is scaled according to the ratio between the average leg length and spacing between two legs (approximately $1.2:1$).
% WRITE ABOUT LEG LENGTHS AND HOW THEY ARE ASSEMBLED TO THE ROBOT
Blue markers are attached to the top side of the robot's links and pink markers on the right side of the links for orientation tracking. 
% The robot was powered by a 12V DC power supply.

\subsection{The Self-righting Gait}

Based on our observations of the wave-based and one-shot righting behaviors of \textit{S.~subspinipes}, we model the self-righting gait of our robophysical platform using a general framework that unifies sidewinding and rolling gaits, both of which are well documented in limbless elongated robots. Specifically, the gait is described as the superposition of a horizontal (yaw) and a vertical (pitch) traveling wave propagating along the body~\cite{astley2015modulation,chong2022general}. Let $\alpha_{y}(i,t)$ and $\alpha_{p}(i,t)$ denote the $i$-th yaw (lateral) and pitch (vertical) joint angles at time $t$, respectively. The resulting body wave can be expressed as:
\begin{align}\label{eq:sidewind}
    & \alpha_{y}(t,i) = A_{y}\sin{(\omega t + 2\pi i\frac{n_{y}}{N} + \Delta d)} \nonumber \\
    & \alpha_{p}(t,i) = A_{p}\sin{(\omega t + 2\pi i\frac{n_{p}}{N})}
\end{align}

\noindent where $A_{y}$ and $A_{p}$ are the yaw and pitch amplitude respectively, $\omega$ is the temporal frequency, $n_{y}$ and $n_{p}$ are the number of horizontal and vertical body waves along the robot respectively, $N$ is the number of yaw/pitch joints with $N=9$ for our robophysical model, and $\Delta d$ is the phase offset between the yaw and pitch joints.
Note that when $A_{y}=A_{p}$, $n_{y}=n_{p}=0$ and $\Delta d=-\pi/2$, the sidewinding equations become the equations for rolling:

\begin{align}\label{eq:roll}
    \alpha_{y}(t,i) = A_{y}\sin{\omega t}, \ \ 
    \alpha_{p}(t,i) = A_{p}\cos{\omega t} 
\end{align}

\noindent which is another widely used locomotion gait for elongated robots ~\cite{hatton2010generating}.

\subsection{Experiments}

To determine the feasible self-righting strategies across different robot morphologies, we conduct systematic experiments on flat, rigid ground in our laboratory workspace. Robot motion is recorded using 240~fps cameras mounted above and to the side of the testing area (see Fig.~\ref{fig:fig4}(A)). 

Across experiments, we observe three distinct behaviors: 
(1) \textit{Sidewinding}, where the robot translates laterally without significant axial rotation or self-righting (Fig.~\ref{fig:fig4}B.1); 
(2) \textit{In-place self-righting}, where the robot exhibits substantial axial body roll while remaining approximately stationary (Fig.~\ref{fig:fig4}B.2); and 
(3) \textit{Sidewinding spin}, in which lateral translation and axial rotation occur simultaneously (Fig.~\ref{fig:fig4}B.3). 

While the first two behaviors are well documented in limbless locomotion, to the best of our knowledge this is the first observation of a locomotion mode in which axial rotation about the body helix is coupled with net lateral translation. 
This translation–rotation coupling may enable elongated robots to encircle cylindrical structures while advancing, potentially increasing effective payload capacity. Accordingly, we quantified both lateral displacement (body lengths traveled per cycle, BL/cycle) and axial rotation (average body spin per cycle, rad/cycle) to characterize the full behavior space. We record five trials 
% {\color{red}GB: trials?}
for each combination of parameters and recorded the displacement and axial rotation for each trial. 
% {\color{red}GB: trial? There are more instances of trail below.} . 

\subsection{Behavior Diagram and parameter space}

\begin{figure}[t]
    \centering
    \includegraphics[width=1.0\linewidth]{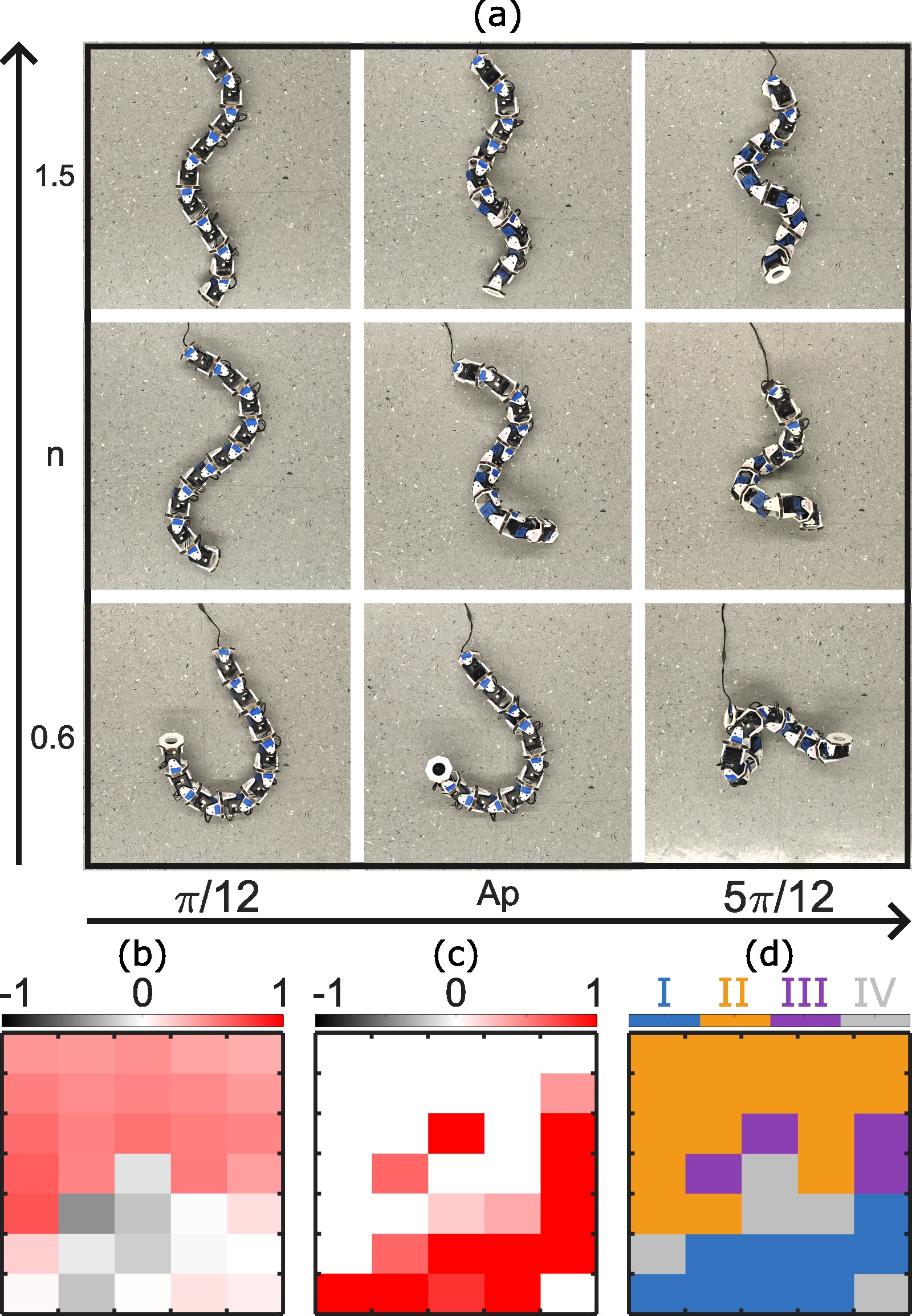}
    \caption{\textbf{Behavior diagrams of displacement and axial rotation.} 
    (a) Parameter space of limbless robot self-righting, with axes corresponding to pitch-joint amplitude ($A_p$) and number of body waves ($n$). 
    (b) Behavior diagram of net displacement over the parameter space (color bar unit: body lengths per cycle, BL/cycle). 
    (c) Behavior diagram of average axial rotation over the parameter space (color bar unit: rad/s).
    (d) Locomotion mode distribution across the parameter space. I: self-righting in place; II: sidewinding; III: sidewinding spinning; IV: kinematic saturation.}
    \label{fig:fig6}
    \vspace{-2em}
\end{figure}

We construct a behavior diagram for the displacement and axial rotation respectively for each parameter combination (see Fig.\ref{fig:fig6}B and C)  to visualize the results and snapshots of the limbless robot's configuration across the parameter space of $A_{p}$ and $n$ are shown in Fig.\ref{fig:fig6}A.  Specifically, we fixed $A_{y}=\pi/4$, $\Delta d=\pi/2$ and gradually incremented $A_{p}$ and $n_{y}=n_{p}$ based on the equation (\ref{eq:sidewind}) to investigate the effect of the vertical body wave amplitude and the number of body waves on the locomotion modes, displacement, and axial rotation of our limbless robot. We started from $A_{p}=\pi/12$ and $n_{y}=n_{p}=n=0.6$ and used the resolution of $\pi/12$ for $A_{p}$ and 0.15 for $n$. We record five trials for each combination of parameters and record the displacement and axial rotation for each trial. 

Finally, we fix short, medium, and long legs on the bottom of each yaw joint and repeat the trials across the parameter space of $A_{p}$ and $n$ for each set of legs to investigate the effect of leg length on both displacement and self-righting success rate and determine feasible self-righting strategies for each robot morphology (see Fig.\ref{fig:fig7}). We also repeat the trials with different number of legs mounted on the robot to test how the leg number affects the displacement, self-righting success rate, and the parameter combinations that form feasible self-righting strategies.

\begin{figure}
    \centering
    \includegraphics[width=\linewidth]{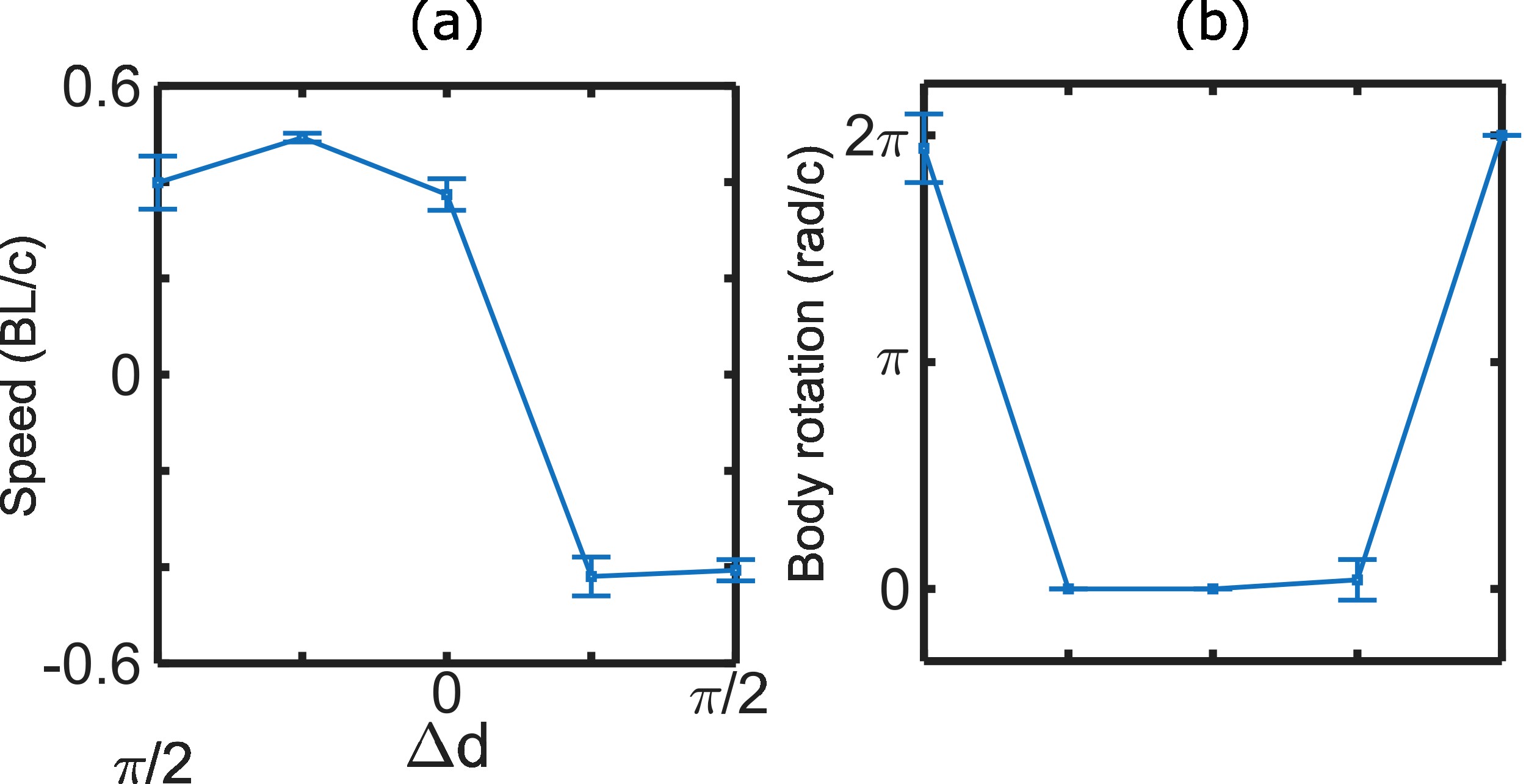}
    \caption{\textbf{Robot Displacements under different yaw and pitch wave phase offsets} (a) Forward displacement of the limbless robot in body lengths (BL) per cycle as a function of \(\Delta d\), the difference between the phase offset between the yaw joints and pitch joints. (b) Robot spinning (axial rotation) per cycle as a function of \(\Delta d\).}
    \label{fig:fig5}
\end{figure}%

\begin{figure}
    \centering
    \includegraphics[width=\linewidth]{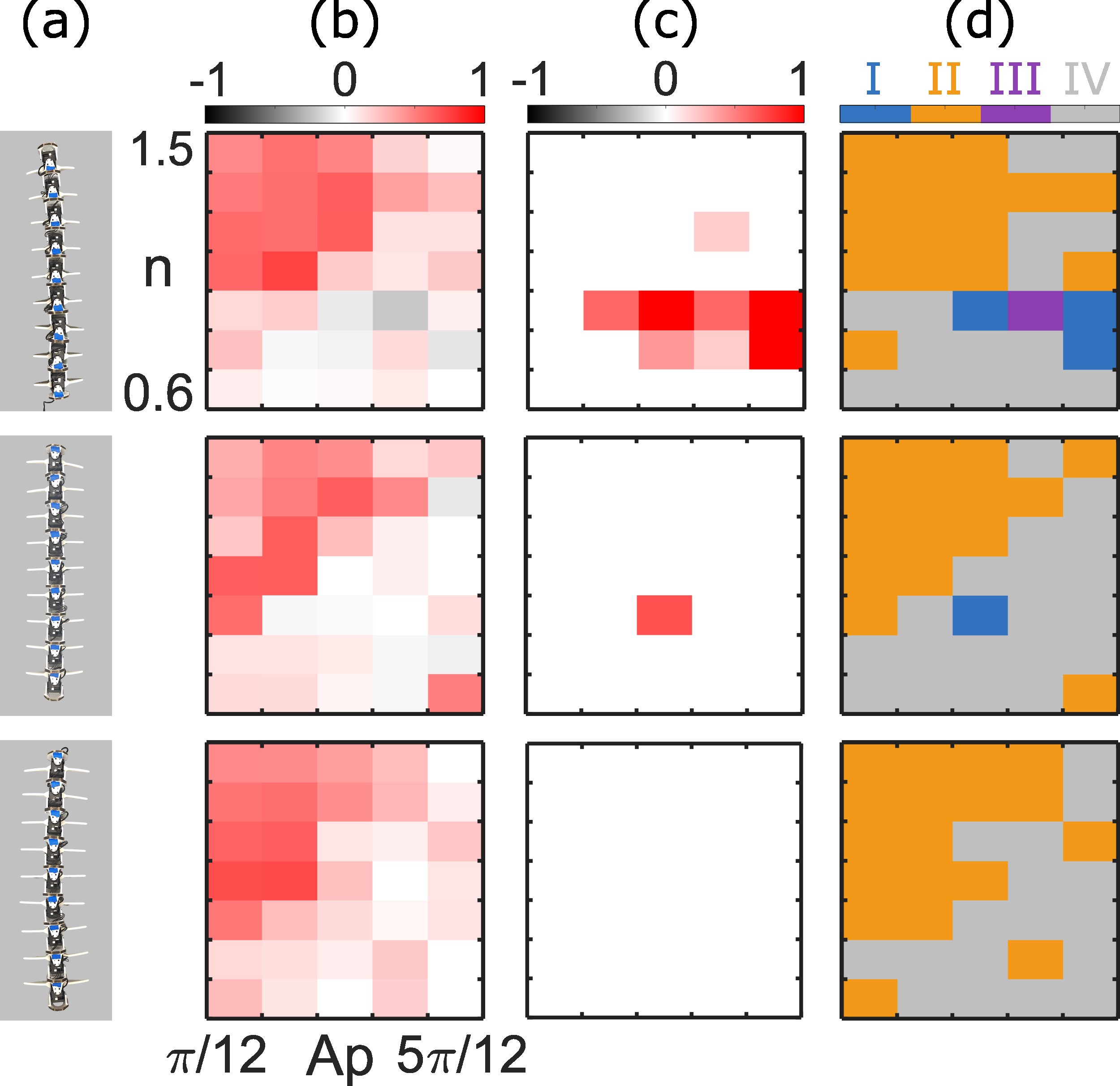}
    \caption{\textbf{Effect of limb size on behavior diagrams.} 
    (a) From top to bottom: the robot equipped with nine pairs of short, medium, and long legs. 
    (b) Behavior diagrams of net displacement over the parameter space (color bar unit: BL/cycle). 
    (c) Behavior diagrams of average axial rotation over the parameter space (color bar unit: rad/s). 
    We observe that the successful self-righting rate decreases as limb length increases: only a single parameter combination yields successful self-righting for medium-length legs, and none for long legs.
    (d) Locomotion mode distribution across the parameter space. I: self-righting in place; II: sidewinding; III: sidewinding spinning; IV: kinematic saturation.}
    \label{fig:fig7}
    \vspace{-1em}
\end{figure}%

\section{Results}

\subsection{Effect of Phase Offset on Locomotion Modes}

In prior literature, it was separately documented that the phase offset between pitch and yaw body waves are set to be $\pi/2$ for both sidewinding~\cite{chong2021frequency,astley2015modulation} and for rolling~\cite{choset_ComplexPipe}. 
Within our unified framework, we examine whether this phase relationship is fundamentally required or merely a coincidence. To evaluate this, we systematically vary the phase offset $\Delta d$ from $-\pi/2$ to $\pi/2$ in increments of $\pi/4$, and quantified the resulting lateral displacement (characterizing sidewinding performance; unit: body length per cycle, BL/c) and axial rotation (characterizing rolling and self-righting performance; unit: axial rotational displacement per cycle, rad/c) across the parameter space. We choose the amplitudes to be $\pi/3$, large enough to generate both sidewinding and rolling behaviors.

Surprisingly, we observe that lateral displacement is relatively insensitive to the phase offset $\Delta d$: no statistically significant differences in translational absolute speed are observed across the tested values of $\Delta d$. In contrast, as shown in Fig.~\ref{fig:fig5}, axial rotation is highly sensitive to $\Delta d$. A full body revolution is achieved only when $\Delta d = \pm\pi/2$, whereas negligible axial rotation occurs for other phase offsets.

%To this end, we argue that the distinction between sidewinding and sidewinding spin is governed by the choice of phase offset $\Delta d$. The robot exhibits positive lateral displacement when $\Delta d < 0$ and negative displacement when $\Delta d > 0$. When $\Delta d = 0$, the direction of motion depends sensitively on the robot’s initial configuration, resulting in unstable and highly variable behavior (as evidenced by large standard deviations).

Based on this,  
% {\color{red}GB:Based on this, instead of To this end?}
we argue that the distinction between sidewinding and sidewinding spin is governed by the choice of phase offset $\Delta d$. Importantly, the underlying mechanisms of lateral displacement differ across the phase space. When $\left| \Delta d\right| - \frac{\pi}{2} $ is large, the robot remains largely statically stable during locomotion, and propulsion is primarily generated through active lifting and placement of body segments. In contrast, when $\left| \Delta d\right| - \frac{\pi}{2} $ is small, the robot becomes statically unstable during locomotion, and net displacement is driven by gravity-assisted rolling dynamics, resembling a persistent self-righting process. 

We posit that ground-wave self-righting observed in biological centipedes shares a similar mechanism with the rolling-dominated sidewinding regime. Naturally, the exact locomotion performance (i.e., translational and rotational displacement) also depends on other gait parameters, such as the amplitudes and frequencies of the pitch and yaw waves. In the following sections, we systematically investigate how self-righting and sidewinding dynamics vary as functions of these gait parameters. To simplify the analysis and isolate their effects, we fix the phase offset at $\Delta d = \pi/2$ unless otherwise noted.

%Both $\Delta d=0$ and $\Delta d=\pi$ generates substantial translational displacement and axial rotation simultaneously, which increases the probability of robot self-righting. The only difference between $\Delta d=0$ and $\Delta d=\pi$ is that the robot travels in opposite direction.
%Therefore, we fix $\Delta d=0$ for the rest of the experiments to stay consistent with other literature.

% ###############################################################

\subsection{Effects of Gait Parameters}
We construct a parameter space for a unified family of sidewinding and self-righting gaits, with two primary axes: the vertical body-wave amplitude $A_{p}$ and the number of body waves $n$ along the body (Fig.~\ref{fig:fig6}a). Notably, robots executing these gaits often operate near the boundary between statically stable and statically unstable configurations. Consequently, locomotion performance is highly sensitive to implementation details, including robot geometry, material compliance from 3D-printed parts, and low-level motor control (e.g., PID tuning). Preliminary simulation studies reveal a substantial simulation-to-reality gap, limiting the reliability of purely simulation-based predictions. To obtain quantitative insights, we experimentally implement these gaits on the physical robot and systematically measure the resulting axial rotation (Fig.~\ref{fig:fig6}c) and lateral displacement (Fig.~\ref{fig:fig6}b) across the parameter space, thereby constructing empirical behavior diagrams.
% Behavior diagrams of displacement per cycle and axial rotation per cycle for the limbless robot are shown in Fig.~\ref{fig:fig7}, and the behavior diagrams for the limbed robot are shown in Fig.~\ref{fig:fig8}.
% An angular displacement (axial rotation) value closer to 1 means the success rate of self-righting approaches 1 since a feasible self-righting strategy for that particular robot morphology will rotate the robot in a full revolution around its body axis within one cycle for the limbless robots. Similarly, for the robot with limbs, it will self-right to the upright configuration then flip back to the upside down starting configuration, similar to what has been reported in \cite{erikSR}.

We observe four distinct behavioral regimes across the $(A_{p}, n)$ parameter space:

\begin{enumerate}
    \item \textbf{In-place spinning (one-shot regime).} 
    For low values of $A_{p}$ and $n$, the robot adopts a ``C''-shaped configuration and generates an almost standing body wave, resulting primarily in axial spinning with minimal net translation. 
    As $A_{p}$ increases (while keeping $n$ low), the body progressively forms a helical configuration. 
    In both cases, the robot exhibits in-place self-righting behavior characterized by substantial axial rotation but negligible displacement (see Fig.~\ref{fig:fig4}B.2). 
    These gaits closely resemble the one-shot ground self-righting behaviors observed in animals, relying on rapid, dynamically generated energy to overcome the gravitational energy barrier~\cite{erikSR}.

    \item \textbf{Kinematic saturation regime.} 
    For large $n$ and very high $A_{p}$ (approaching the joint-angle limits of the robophysical platform), neither coherent sidewinding nor axial spinning is observed. 
    Instead, the robot enters a mechanically constrained regime where excessive curvature prevents effective wave propagation.

    \item \textbf{Pure sidewinding regime.} 
    For large $n$ and low $A_{p}$, the robot exhibits lateral sidewinding with reduced or negligible axial rotation. 
    Locomotion in this regime is dominated by translational displacement rather than body rolling.

    \item \textbf{Rolling-assisted sidewinding regime.} 
    For intermediate values of $A_{p}$ at larger $n$, lateral translation is coupled with axial rotation. 
    In this regime, rolling contributes to forward progression, resembling the ground-wave self-righting behavior observed in animals.
\end{enumerate}

In Fig.~\ref{fig:fig7}d, we present a phase diagram over the $(A_{p}, n)$ parameter space, classifying the four behavioral regimes based on empirically defined thresholds in axial rotation and lateral displacement. We define in-place spinning as having lateral displacement $dX<0.3$ BL/cycle and axial rotation $\dot{\theta}>0.5$ rad/s; kinematic saturation as $dX<0.3$ BL/cycle and $\dot{\theta}<0.5$ rad/s; pure sidewinding as $dX>0.3$ BL/cycle and $\dot{\theta}<0.5$ rad/s; and rolling-assisted sidewinding as $dX>0.3$ BL/cycle and $\dot{\theta}>0.5$ rad/s.
%and a larger $n$ for each value of $A_{p}$ reduces the sidewinding displacement, while increasing $A_{p}$ has a lower reduction in displacement in comparison. We also noticed significant steering during sidewinding particularly in the cases where $A_{p}=\pi/12,\pi/6$ and $n<1.05$. The robot was constantly steering while sidewinding, essentially going in a circular path, which explains the negative displacements appeared in Fig.\ref{fig:fig6}(b). Steering during sidewinding has been documented by \cite{chong2021frequency}, where changing the ratio between $n_{y}$ and $n_{p}$ can lead to clockwise turning. In our work, changing the ratio between $A_{y}$ and $A_{p}$ led to counterclockwise turning.

% ###################################################################

\subsection{Effects of Gait Parameters and Leg Length}

For multi-legged elongated robots, we investigate how the phase diagram changes as a function of limb morphology. To this end, we mount pairs of 3D-printed, non-actuated legs of three different lengths onto our robophysical platform. We then repeat the experimental procedure performed for the limbless robot to examine the combined effects of $A_{p}$, $n$, and leg length on self-righting and displacement (see Fig.~\ref{fig:fig7}a). We observe that increasing leg length makes self-righting progressively more challenging, as evidenced by the substantial reduction in the feasible regions corresponding to in-place self-righting (blue) and sidewinding spin (purple) regimes in the phase diagram.

\begin{figure}
    \centering
    \includegraphics[width=\linewidth]{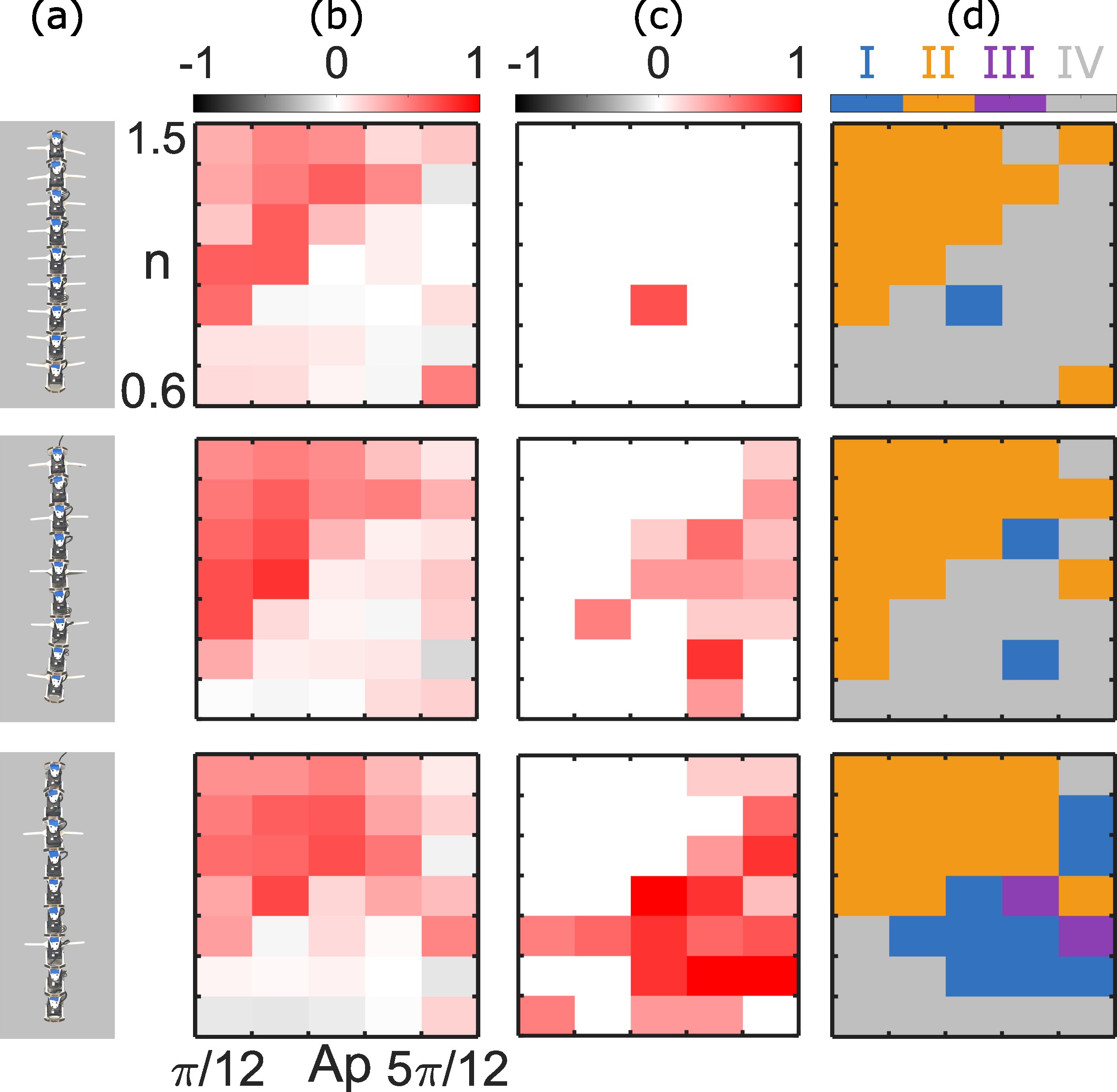}
   \caption{\textbf{Effect of limb number on behavior diagrams.} 
    (a) The robot equipped with different numbers of legs (leg length: 6.06 cm). 
    (b) Net displacement over the parameter space (BL/cycle). 
    (c) Average axial rotation over the parameter space (rad/s). 
    As the number of legs increases, the feasible region for successful self-righting decreases.
    (d) Locomotion mode distribution across the parameter space. I: self-righting in place; II: sidewinding; III: sidewinding spinning; IV: kinematic saturation.}

    \label{fig:fig8}
    \vspace{-2em}
\end{figure}%

The results suggest that the presence of legs introduces an additional energy barrier that the actuators must overcome to achieve successful self-righting, consistent with arguments presented in~\cite{charles_MPCrecover}. In particular, longer legs require the robot’s center of mass (CoM) to be elevated to a greater height during reorientation, thereby increasing the gravitational potential energy required for recovery. Moreover, longer legs increase the likelihood of self-collision, which constrains the feasible range of body postures. As a result, higher wave amplitudes may become kinematically inaccessible due to geometric interference between limbs and body segments, further reducing the region of viable self-righting strategies. As a result, we notice that, subject to our to gait prescription and hardware, we cannot find a reliable self-righting strategy for limb size greater than 1.2 times the the robot's body segment width.

\subsection{Effects of Gait Parameters and Leg Number}

We next generalize our findings to multi-legged robots with different numbers of legs. Specifically, we repeat the experimental protocol for robots equipped with 9, 5, and 2 pairs of legs (Fig.\ref{fig:fig8}). We observe that as the number of legs decreases, the feasible region for self-righting (red-shaded regions in Fig.~\ref{fig:fig8}c) expands significantly. We attribute this trend to a reduced energy barrier for propagating a self-righting wave and overcoming the gravitational potential barrier.

In particular, as the number of legs increases, larger wave numbers $n$ are generally required to achieve successful self-righting. Based on the trends observed in Fig.~\ref{fig:fig7} and Fig.~\ref{fig:fig8}c, we argue that wave-based (higher $n$) righting is energetically more favorable for elongated robots than one-shot self-righting (lower $n$). In the wave-based strategy, only a portion of the body is elevated at any given time, whereas the one-shot strategy requires simultaneous elevation of the entire body. Consequently, wave-based righting reduces the instantaneous energy and torque demands on the actuators, leading to higher self-righting success rates and a broader range of feasible gait parameter combinations.

Finally, we note that the introduction of legs increases the energetic barrier to axial rolling, which in turn stabilizes sidewinding behaviors. In the limbless configuration, certain gait parameters produce a spinning–sidewinding regime characterized by coupled axial rotation and translation. With the addition of legs, axial rolling is suppressed, effectively biasing the dynamics toward pure sidewinding. As a result, the multi-legged elongated robot achieves substantially higher lateral displacement. Specifically, sidewinding in the multi-legged configuration reaches speeds up to 0.8~BL/cycle when $A_{y}=\pi/4$, $A_{p}=\pi/6$, and $n=1.05$, significantly outperforming the previously reported best sidewinding performance for elongated limbless robots (approximately 0.4~BL/cycle) as well as the best forward gait for multi-legged elongated robots (approximately 0.45~BL/cycle).

\section{CONCLUSION AND FUTURE WORK}

In this study, we compare the self-righting behaviors of \textit{Scolopendra subspinipes} (short legs) and \textit{Scutigera coleoptrata} (house centipede, long legs). We observe that \textit{S.~subspinipes} reliably self-rights both during aerial phases (analogous to cat-like reorientation) and through ground-assisted maneuvers, whereas \textit{S.~coleoptrata} predominantly relies  on aerial reorientation and exhibits difficulty generating effective ground-reaction torques for self-righting. Motivated by these observations, we construct a parameterized space of bio-inspired self-righting strategies defined by body-wave frequency and amplitude, and develop an elongated robotic platform with adjustable leg lengths. Systematic experiments reveal that increasing leg length necessitates a shift in control strategy—specifically, larger wave amplitudes and higher wave numbers—to overcome the increased energetic barrier introduced by the legs. Finally, we find that although legs increase the difficulty of self-righting, they simultaneously stabilize sidewinding behaviors. This stabilization suppresses undesired axial rolling and enables pure sidewinding with speeds up to 0.8~BL/cycle—nearly twice the previously reported maximum sidewinding speed for comparable elongated robots.

% XXX elaborate this!
% 1. In future work, we will build model and seek to make predictions on gaits for self-righting, and optimize for fastest or most energestically efficient gait.
% 2. In future work, we will seek whether there exist an upper bound of leg size beyond which self-righting is not feasible above this point (subject to the power density.
% 3. In future work, we will study how legs can stabilize sideewinding and quantify it as a function of terrain complexity; and comare the sidewinding multi-legged robots VS forward-moving multi-legged robots.

In future work, we will work towards a mathematical model to facilitate the prediction of the optimal self-righting gaits under diverse circumstances.
This mathematical model will also provide guidelines for optimizing the self-righting gaits for metrics including speed and energy efficiency for different robot morphologies.
We will investigate whether a fundamental upper bound exists on leg length beyond which self-righting becomes infeasible due to constraints in power density. 
% Future work will systematically vary leg geometry while holding actuator power density fixed to empirically determine this upper bound. 
Identifying this threshold will provide important scaling laws that inform the design of elongated multi-legged robots across size regimes.
While previous work \cite{erikSR} has revealed that legs can stabilize and improve the sidewinding performance, we will further investigate the fundamental princicples of the effect of legs on sidewinding and quantify it in terms of terrain complexity.
We will also conduct a systematic comparative study between sidewinding elongated multi-legged robots and those that execute forward moving gaits.
We will analyze the trade-offs of these two gaits and their respective advantages in different conditions, potentially discovering fundamental principles of gait selection for elongated multi-legged robots under various circumstances.

% \section*{ACKNOWLEDGMENTS}

% The authors would like to thank Esteban Flores for his help running experiments and his discussion and Chris Pierce for his image analysis expertise. They would also like to thank the Georgia Tech Physics REU and the National Science Foundation for their logistical, educational, and financial support (NSF Grant 2244423). The authors would also like to thank the Institute for Robotics and Intelligent Machines at Georgia Tech for the use of experimental space. The authors received funding from NSF-Simons Southeast Center for Mathematics and Biology (Simons Foundation SFARI 594594), the Army Research Office grant W911NF-11-1-0514, and a Dunn Family Professorship. 

\addtolength{\textheight}{-12cm}   % This command serves to balance the column lengths
                                  % on the last page of the document manually. It shortens
                                  % the textheight of the last page by a suitable amount.
                                  % This command does not take effect until the next page
                                  % so it should come on the page before the last. Make
                                  % sure that you do not shorten the textheight too much.

%%%%%%%%%%%%%%%%%%%%%%%%%%%%%%%%%%%%%%%%%%%%%%%%%%%%%%%%%%%%%%%%%%%%%%%%%%%%%%%%

%%%%%%%%%%%%%%%%%%%%%%%%%%%%%%%%%%%%%%%%%%%%%%%%%%%%%%%%%%%%%%%%%%%%%%%%%%%%%%%%

%%%%%%%%%%%%%%%%%%%%%%%%%%%%%%%%%%%%%%%%%%%%%%%%%%%%%%%%%%%%%%%%%%%%%%%%%%%%%%%%

%%%%%%%%%%%%%%%%%%%%%%%%%%%%%%%%%%%%%%%%%%%%%%%%%%%%%%%%%%%%%%%%%%%%%%%%%%%%%%%%

\bibliographystyle{unsrt}
\bibliography{IROS_bib}

\end{document}